\documentclass[letterpaper, 10pt, conference]{ieeeconf}      

\IEEEoverridecommandlockouts                              

\usepackage{graphicx}
\usepackage[fleqn]{amsmath}
\usepackage[varg]{txfonts}
\usepackage{xcolor}
\usepackage{enumerate}

\title{\LARGE \bf
Material Classification\\ Using Active Temperature Controllable Robotic Gripper
}

\author{Yukiko Osawa$^{1*}$,~\IEEEmembership{Member,~IEEE,} Kei Kase$^{2*}$,~\IEEEmembership{Student member,~IEEE,} Yukiyasu Domae$^{3}$,~\IEEEmembership{Member,~IEEE,}\\ Yoshiyuki  Furukawa$^{4}$, and Abderrahmane Kheddar$^{5}$,~\IEEEmembership{Senior Member,~IEEE}
\thanks{*This work was supported by KAKENHI Grant-in-Aid for Young Scientists Number 21K21325.}
\thanks{$^{1}$National Institute of Advanced Industrial Science and Technology (AIST), Industrial Cyber-Physical Systems Research Center, Tokyo, Japan
        {\tt\small yukiko.osawa-akiyama@aist.go.jp}}
\thanks{$^{2}$National Institute of Advanced Industrial Science and Technology (AIST), Artificial Intelligence Research Center, Tokyo, Japan
        {\tt\small kase@idr.ias.sci.waseda.ac.jp}}
\thanks{$^{3}$National Institute of Advanced Industrial Science and Technology (AIST), Artificial Intelligence Research Center, Tokyo, Japan
        {\tt\small domae.yukiyasu@aist.go.jp}}
\thanks{$^{4}$National Institute of Advanced Industrial Science and Technology (AIST), Industrial Cyber-Physical Systems Research Center, Tokyo, Japan
        {\tt\small y-furukawa@aist.go.jp}}
\thanks{$^{5}$CNRS-University of Montpellier, LIRMM, Interactive Digital Humans group, Montpellier, France
        {\tt\small kheddar@lirmm.fr}}
}

\makeatletter
\newcommand{\figcaption}[1]{\def\@captype{figure}\caption{#1}}
\newcommand{\tblcaption}[1]{\def\@captype{table}\caption{#1}}
\makeatother

\begin{document}

\maketitle
\thispagestyle{empty}
\pagestyle{empty}

\begin{abstract}
Recognition techniques allow robots to make proper planning and control strategies to manipulate various objects. Object recognition is more reliable when made by combining several percepts, e.g., vision and haptics. One of the distinguishing features of each object's material is its heat properties, and classification can exploit heat transfer, similarly to human thermal sensation. Thermal-based recognition has the advantage of obtaining contact surface information in real-time by simply capturing temperature change using a tiny and cheap sensor. However, heat transfer between a robot surface and a contact object is strongly affected by the initial temperature and environmental conditions. A given object's material cannot be recognized when its temperature is the same as the robotic grippertip. We present a material classification system using active temperature controllable robotic gripper to induce heat flow. Subsequently, our system can recognize materials independently from their ambient temperature. The robotic gripper surface can be regulated to any temperature that differentiates it from the touched object's surface. We conducted some experiments by integrating the temperature control system with the Academic SCARA Robot, classifying them based on a long short-term memory (LSTM) using temperature data obtained from grasping target objects.
\end{abstract}

\begin{figure}[t!]
\begin{center}
\includegraphics[width=\columnwidth]{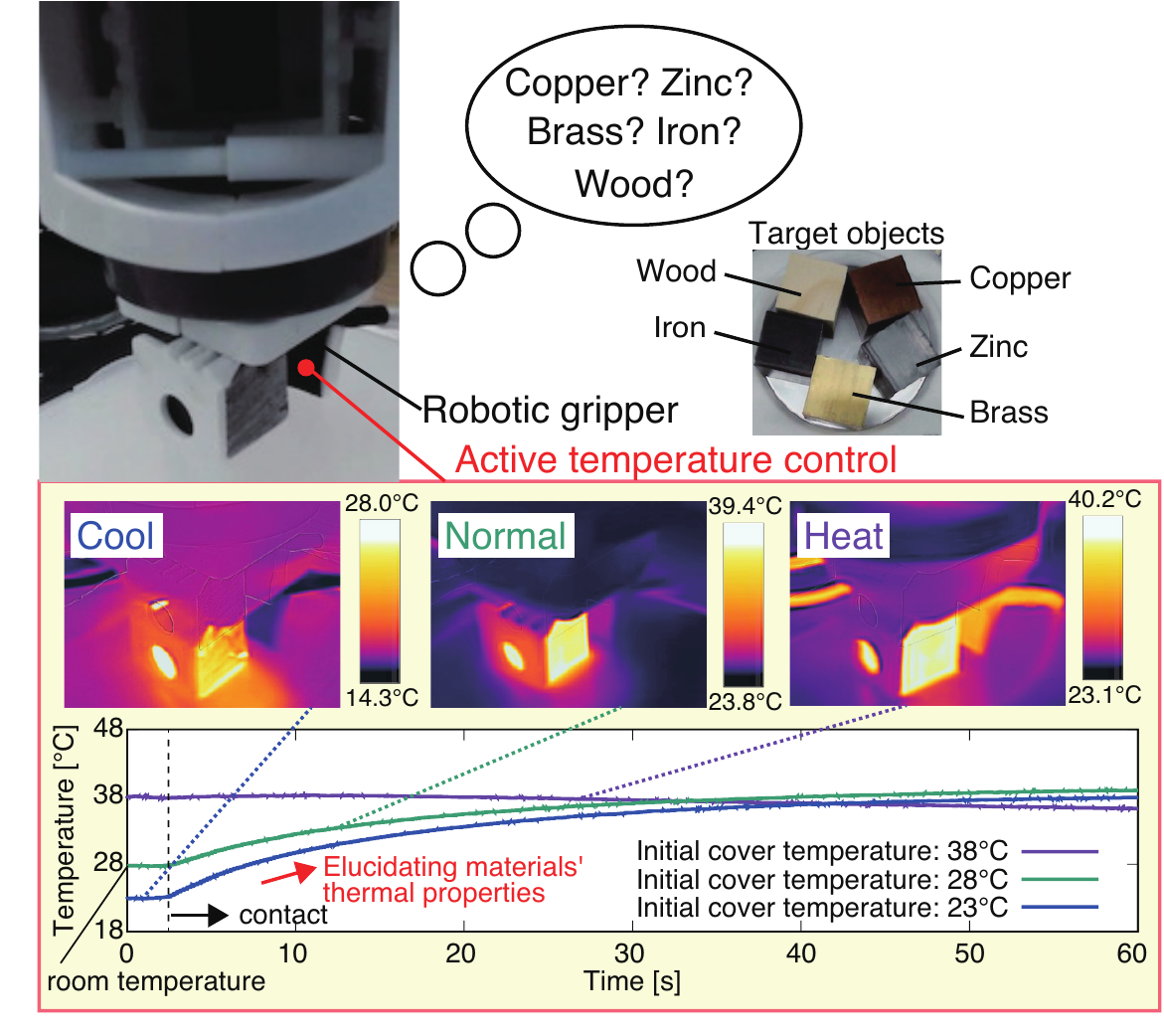}
\caption{The key components of our system.}
\label{fig:concept}
\end{center}
\end{figure}

\section{Introduction}
Sensing and recognizing the surroundings is paramount for robots to perform complex and high-quality tasks.
Especially, contact-based recognition techniques of objects enrich robot perception capability in achieving physical interaction tasks. Robotic perception accuracy and reliability are indeed enhanced by combining vision and haptics~\cite{yuan2018icra,takahashi2019icra,fujita2020ar,inami2021sii}.

One of the distinguishing features of each object's material is heat properties. When two distinct objects having different temperatures are in contact, their contact area gradually converges to a common temperature. Note that this contact temperature depends on each material’s initial temperature and ``thermal effusivity''. In particular, thermal effusivity determines which material's initial temperatures affect the contact surface. This thermal behavior and parameters play an important role in recognizing a given material from temperature responses.

The relationship between the thermal effusivity and the temperature change have been investigated in~\cite{siegel1986icra,russell1988robotica,monkman1993ra}. Since the parameter also relates to human thermal sensation, it plays a role in reproducing the thermal properties of different materials in the telepresence and the haptics research fields. Indeed, several studies proposed thermal rendering using a Peltier device by model-based estimation, e.g.~\cite{yamamoto2004icra,jones2009perception,ho2018temperature}. In~\cite{guiatni2009presence} recorded heat flux obtained from several materials are classified based on neural network learning technique for haptics rendering.

Thermal-based recognition has the advantage of obtaining contact surface information in real-time simply by capturing temperature change using a tiny and cheap sensor. Therefore, these techniques combined with a robotic system have been researched for many years. For example, the studies in~\cite{engel2006sensors,takamuku2008iros,katoh2009ssam} have developed the temperature and heat flow sensing system attached to the actuator, classified contact objects using the obtained signals. In addition, some studies use the thermistor embedded in the ${\rm BioTAC^{TM}}$ biomimetic tactile sensor~\cite{fishel2009robio}. The latter proposed a classification based on Bayesian exploration~\cite{xu2013icra}. Some studies also used the ${\rm BioTAC^{TM}}$'s signal with machine learning such as Artificial Neural Network (ANN)~\cite{kerr2013robio} and Hidden Markov Models (HMMs)~\cite{chu2015ras}. Rapid recognition estimation using learning technique is proposed in~\cite{bhattacharjee2015rss}, extended it to the binary classification in real life~\cite{bhattacharjee2016haptics}, and developed the multimodal sensors~\cite{bhattacharjee2018ral}. 
 
In fact, thermal response depends on the initial temperature of the robotic fingers' surface and the contacted object~\cite{bhattacharjee2015rss,bhattacharjee2020arxiv}. Notably, most conventional studies cannot actively control the surface temperature. Hence, a given object cannot be recognized when its temperature is the same as the robotic contacting surface, and the environmental change reduces classification accuracy (e.g., temperature of an object could change through the industrial machining process and room temperature). 

We tackle this shortcoming using an active temperature control system integrated to the robot gripper. This is made thanks to the temperature-controllable robotic cover~\cite{osawa2021scirep,osawa2020iser}. Note that our proposed method also aims to be applied to human contact recognition in future work; the surface temperature needs to be regulated according to a human body temperature. Figure~\ref{fig:concept} shows the key components of our system; our method classifies materials (e.g., copper, brass, zinc, iron, and wood) in various temperatures by grasping objects of interest. Since the gripper surface can be regulated to any temperature (e.g., the gripper was controlled to 23~$^\circ$C, 28~$^\circ$C, and 38~$^\circ$C when it grasped the heated iron block (43~$^\circ$C) in Fig.~\ref{fig:concept}), it can actively induce heat flow to elucidate materials' thermal properties. The obtained temperature data are used for material classification based on a long short-term memory (LSTM). We conducted some experiments to assess the accuracy of the material classification in various temperatures of the gripper and the objects.

\section{Thermal contact interaction model}
The section explains the thermal interaction model between two objects during contact. In this paper, each object is considered as a ``semi-infinite solid model'', idealizing as an infinity body extending from a single plane surface. 
Thus, the thermal property of an object can be expressed as
\begin{equation}
\frac{\partial T}{\partial t}=\alpha \frac{\partial^2 T}{\partial x^2}, \label{eq:heat_transfer}
\end{equation}
where $T$, $t$, $x$, and $\alpha$ stand for temperature, time, position, and heat diffusivity, respectively. 
The initial and boundary conditions are 
\begin{equation}
T(x,0)=T_i \;\;\; \text{and} \;\;\; T(0,t)=T_s, \label{eq:boundary}
\end{equation}
where $T_i$ and $T_s$ present initial temperature and surface temperature, respectively. 
Using \eqref{eq:heat_transfer}--\eqref{eq:boundary}, temperature of the object is derived as
\begin{equation}
\frac{T(x,t)-T_s}{T_i-T_s} = \frac{2}{\sqrt{\pi}} \int_{0}^{\frac{x}{2\sqrt{\alpha t}}} e^{-\eta^2}d\eta ,
\end{equation}
where $\pi$ denotes circle ratio. 
Thus, temperature $T (x,t)$ can be expressed as
\begin{equation}
T(x,t)=T_s + \left( T_i - T_s \right)\text{erf}\frac{x}{2\sqrt{\alpha t}}, \label{eq:Txt}
\end{equation}
where erf denotes the Gauss error function. 

Following this model, two objects in contact are simplified as two semi-infinite solids. These two objects are regarded as a contact material (subscript $m)$ and a robotic device (subscript $\textit{dev}$) in this study. Since this paper investigate the device's initial temperature on material classification, thermal contact resistance is not considered to simplify the model. 
The thermal transfer between the contact material and the device is conserved as
\begin{equation}
-\lambda_m \frac{\partial T_m}{\partial x_m} = \lambda_{dev} \frac{\partial T_{dev}}{\partial x_{dev}}, \label{eq:balance}
\end{equation}
where subscript $m$, $dev$, and $\lambda$ stand for a parameter of material, device, and heat conductivity, respectively. 
Using \eqref{eq:Txt} and \eqref{eq:balance}, the thermal exchange can be expressed as 
\begin{equation}
-\left( T_{mi}-T_s \right)\frac{\lambda_m}{2\sqrt{\alpha_mt}}=\left( T_{devi}-T_s \right)\frac{\lambda_{dev}}{2\sqrt{\alpha_{dev}t}},
\end{equation}
where $T_{mi}$ and $T_{devi}$ present initial temperature of the material and the device. 

\begin{figure}[t!]
\begin{center}
\includegraphics[width=\columnwidth]{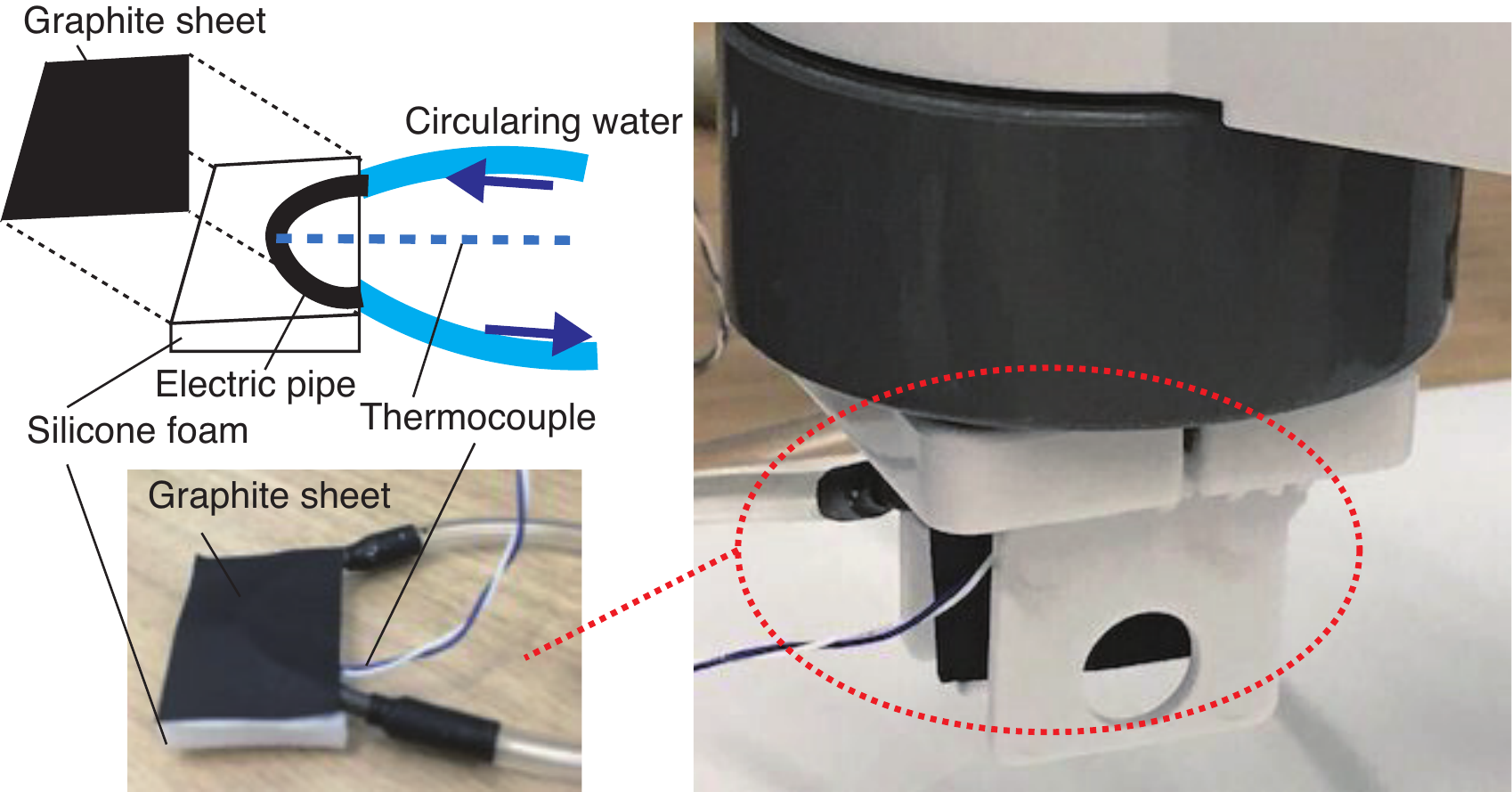}
\caption{The robotic gripper design.}
\label{fig:cover_design}
\end{center}
\end{figure}

Therefore, the contact surface temperature can be calculated as 
\begin{align}
T_s &= \frac{T_{mi} + T_{devi}\gamma}{1 + \gamma} \label{eq:thermal_effusivity}\\
\gamma &= \frac{\sqrt{\lambda_{dev} \rho_{dev} c_{dev}}}{\sqrt{\lambda_{m} \rho_{m} c_{m}}} = \frac{T_s - T_{mi}}{T_{devi} - T_s}, \label{eq:gamma}
\end{align} 
where $\rho$ and $c$ stand for density and specific heat, respectively. 
Here, $\sqrt{\lambda \rho c}$ stands for ``thermal effusivity'', which is called also ``contact coefficient"~\cite{jones2009perception,ho2018temperature}; it combines thermal conductivity, density, and heat capacity, and is a measure of material ability to exchange thermal energy with its surroundings. The equation~$\eqref{eq:thermal_effusivity}$ show that the contact temperature depends on the initial temperature and the thermal effusivity of the material and the device.

In the proposed system, $T_{devi}$ can be regulated to have temperature difference with the material ($\Delta T_{md}$) as
\begin{equation}
T_{devi} = T_{mi} \pm \Delta T_{md}. \label{eq:T_def}
\end{equation}
Thus, the contact temperature can be written using $\Delta T_{md}$ as
\begin{equation}
T_s = T_{devi} \mp \frac{\Delta T_{md}}{1+\gamma}. \label{eq:DeltaT}
\end{equation}
According to \eqref{eq:DeltaT}, the surface temperature does 
not change when $T_{mi}$ and $T_{devi}$ are at the same temperature ($\Delta T_{md}$ equals to zero) since $T_s = T_{mi} = T_{devi}$.

\begin{figure}[t!]
\begin{center}
\includegraphics[width=.9\columnwidth]{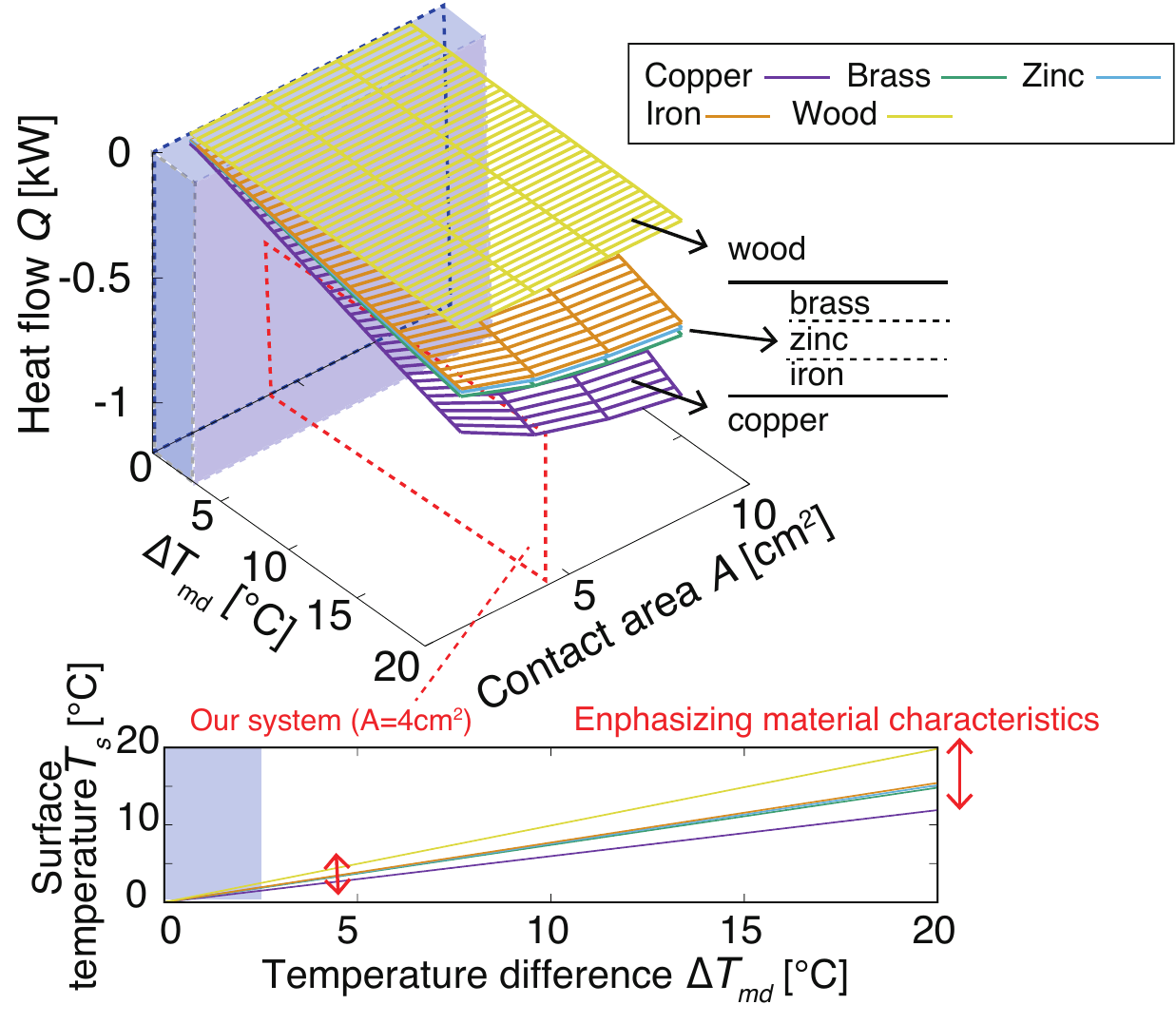}
\caption{Thermal exchange with regard to $\Delta T_{md}$.}
\label{fig:3D_plot}
\end{center}
\end{figure}

	\begin{table}[t!]
		\begin{center}
		\caption{Thermal parameters of the contact objects.} \label{tab:thermal_effusivity}
		\vspace{5pt}
		\begin{tabular}{c|c|c|c}\hline 
		       Materials & $\lambda$ [W/mK] &$\sqrt{\lambda \rho c}$ [$J/m^2s^{1/2}K$] &reference \\
			\hline \hline
		       Copper &386 &3.64$\times 10^{4}$ &\cite{holman1990mcgraw}\\
		       Zinc &112.2 &1.76$\times 10^{4}$ &\cite{holman1990mcgraw}\\
		       Brass &111 &1.91$\times 10^{4}$ &\cite{holman1990mcgraw}\\
		       Iron &73 & 1.61$\times 10^{4}$ &\cite{holman1990mcgraw}\\
		       Wood &0.112 & 3.67$\times 10^{2}$ &\cite{ho2018temperature}\\
			\hline 
		\end{tabular}
		\end{center}
		\end{table}

Using \eqref{eq:Txt}, the derivative of temperature with respect to space ($x$) is calculated as
\begin{equation}
\frac{\partial T(x,t)}{\partial x} = \frac{T_i - T_s}{\sqrt{\pi \alpha t}}e^{-\frac{x^2}{4\alpha t}}. \label{eq:dTdx}
\end{equation}
Therefore, the heat flow through a contact surface ($x=0$) is derived based on~\eqref{eq:DeltaT} and \eqref{eq:dTdx} as
\begin{align}
Q_{dev}&=-\lambda A\frac{\partial T_{dev}}{\partial x} \nonumber = \frac{\lambda A \left(T_s-T_{devi} \right)}{\sqrt{\pi \alpha t}} \nonumber \\
&=\frac{\lambda A \left(\mp \frac{\Delta T_{md}}{1+\gamma} \right)}{\sqrt{\pi \alpha t}}, \label{eq:q_dT}
\end{align}
where $A$ is the contact surface. The simulated heat flowing into contact objects from the heated device ($t=10$~sec) and the device surface temperature are shown in Fig.~\ref{fig:3D_plot}, derived from \eqref{eq:DeltaT} and \eqref{eq:q_dT}. These graphs indicate that the heat movement quantity increases progressively according to increased $\Delta T_{md}$ and $A$, clarifying the material characteristics. Note that distinguishing among metals (copper, brass, zinc, iron...) is more challenging than distinguishing metals from wood. Indeed, metals have close thermal effusivity (see Table~\ref{tab:thermal_effusivity}). Since $\Delta T_{md}$ is controllable, our system can intendedly induce large temperature change depending on contact materials. Besides, actual temperature responses may vary due to slight changes in contact conditions. Repetitive experiments (5 times) with each material (copper, iron, wood), revealed standard deviation of 0.3-0.7~$^\circ$C in temperature responses. Consequently, materials are hard to be distinguished with small $\Delta T_{md}$ (temperature variation can affect classification in the blue area of Fig.~\ref{fig:3D_plot}). Based on the theory, we investigate, using our proposed system, the accuracy of the material classification in function of various $\Delta T_{md}$ and $T_{mi}$. 
\begin{figure}[t!]
\begin{center}
\includegraphics[width=0.8\columnwidth]{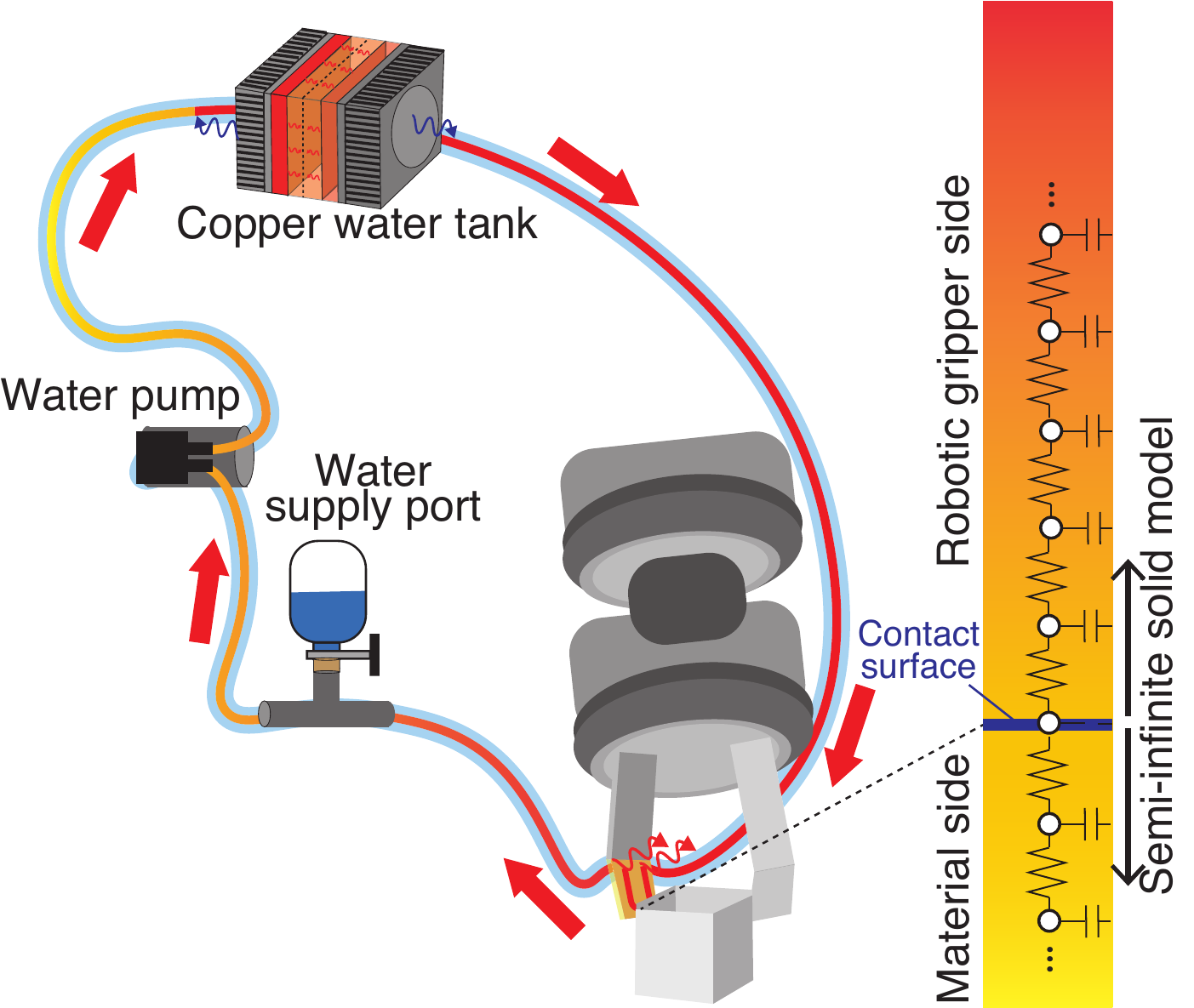}
\caption{Active temperature control system.}
\label{fig:circulating_watersystem}
\end{center}
\end{figure}

\begin{figure*}[t!]
\begin{center}
\includegraphics[width=1.8\columnwidth]{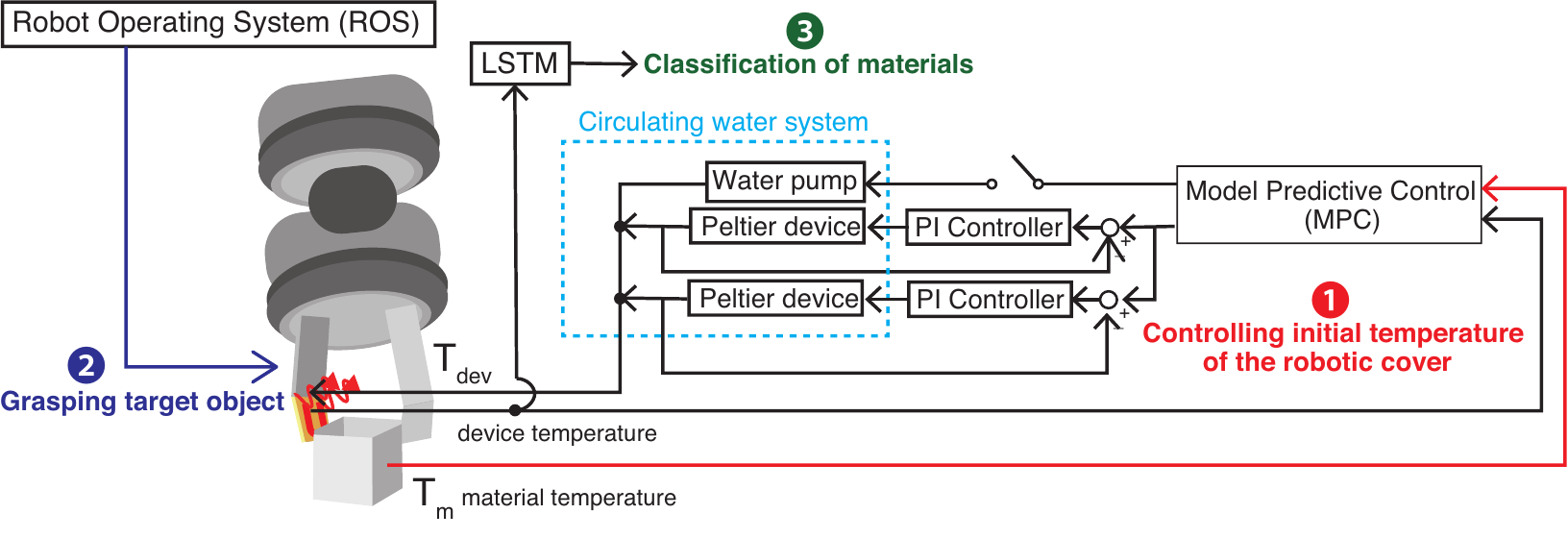}
\caption{Global integration scheme for material recognition using active temperature control system.}
\label{fig:whole_control_system}
\end{center}
\end{figure*}

\begin{figure}[t!]
\begin{center}
\includegraphics[width=\columnwidth]{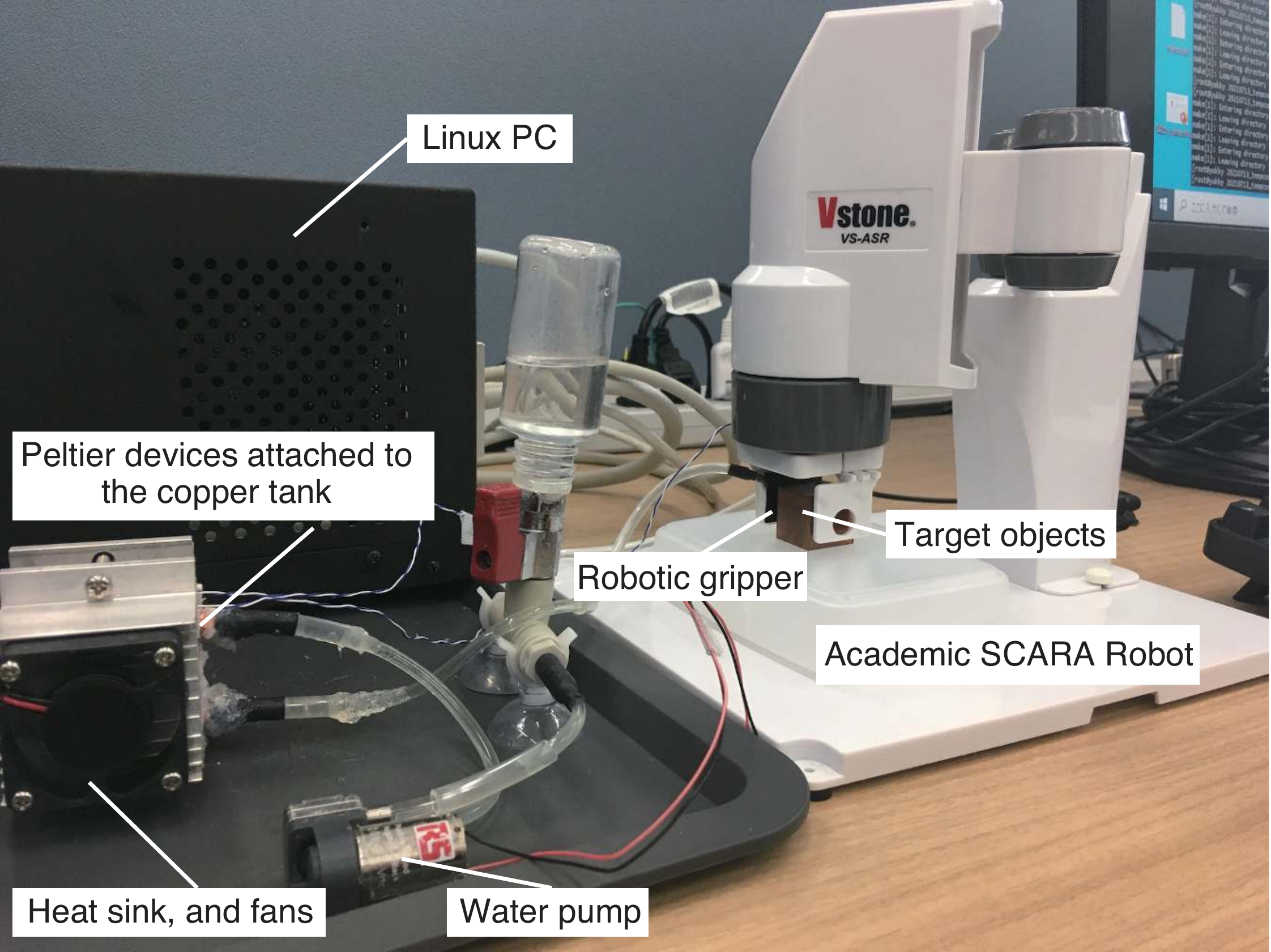}
\caption{Experimental setup.}
\label{fig:ex_setup}
\end{center}
\end{figure}

\section{Active temperature control system\\ integrated with the robotic system}

\subsection{Active temperature control system}
The prototype of the temperature controllable robotic gripper is shown in Fig.~\ref{fig:cover_design}. The basic design borrows from our previous work~\cite{osawa2021scirep,osawa2020iser}. An electric pipe made of Nickel graphite is embedded in a silicone foam, and a thermocouple is attached to the pipe.
The prototype's surface is covered by a graphite sheet of which heat conductivity is high (900~W/mK), avoiding a direct contact between the thermocouple and objects of interest. The temperature of the gripper surface is controlled by circulating temperature-controlled water within the gripper. We used water for temperature control because of the similarity in thermal property with the semi-infinite solid model mentioned in Section~II (see Fig.~\ref{fig:circulating_watersystem}). Thanks to water's high thermal capacity, the system's heat source does not affect thermal exchange at the contact surface during contact. Our solution emulates the human blood system.

The circulating water system consists of the water pump, the water supply port, copper water tank, and robotic gripper. Two Peltier devices are attached to both sides of the copper tank, playing a role in generating and absorbing heat in the system based on a Peltier effect. The temperature of the Peltier device is controlled based on a Model Predictive Control (MPC), using the identified parameters obtained from pre-experiments. 

\subsection{System integration}
Our proposed system consists of the following three parts: $\textcircled{\scriptsize 1}$ the circulating water system for regulating the gripper's temperature, $\textcircled{\scriptsize 2}$ the robotic system for grasping the target objects, and $\textcircled{\scriptsize 3}$ material classification using the gripper's temperature responses based on the long short-term memory (LSTM) neural network (see Fig.~\ref{fig:whole_control_system}).

The temperature responses for material classification are obtained as follows.
First, the temperature command is decided considering the object's temperature and $\Delta T_{md}$, and the gripper surface temperature is controlled to the command (the controlled surface temperature is regarded as $T_{devi}$). The object's temperature is known (ambient temperature in practice, otherwise a step is added to acquire it first to adjust the grippertips temperature). Second, the water pump turns off while contacting the object so that the temperature of the contact surface changes according to the semi-infinite solid model. Then the robot is controlled to grasp the object for 10 sec. Our system can not directly obtain $T_s$; however, the gripper surface temperature is affected from $T_s$, and we used the temperature response for material classification. We are using the Robot Operating System (ROS) for communication and control. The obtained temperature data is processed by the LSTM neural network, implemented in Python.

\section{Experiments}
\subsection{Outline of the experiments}
The soft gripper is attached to the one-side grippertip of the Vstone's Academic SCARA robot (VS-ASR: Lot No. 2121015). Figure~\ref{fig:ex_setup} shows the experimental setup for material classification using the active temperature control system. We use three kinds of $2\times 2$~cm-size blocks (copper, iron, brass, zinc, and wood) as target objects for material classification (Their thermal properties are summarized in Table~\ref{tab:thermal_effusivity}). The objects are heated to 43~$^\circ$C and cooled to 18~$^\circ$C using the cup warmer/cooler (GH-USB-CUP2W) in advance. We conducted three experiments to assess the proposed classification system. As for LSTM, the training/test data (10~sec) is processed by adding Gaussian noise and shifting the initial temperature to augment the data by 100 times. The batch size and epoch number are set to 100 and 2000, respectively. The dataset size is 500 (5 material classification) and 300 (3 material classification). 

\begin{figure}[t!]
\begin{center}
\includegraphics[width=\columnwidth]{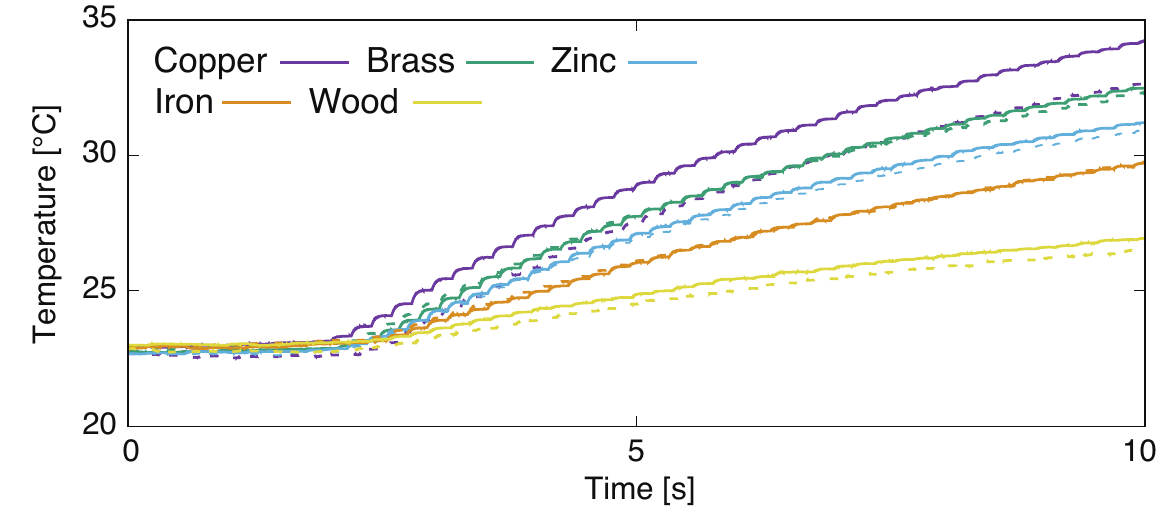}
\caption{[1-A] Test and training data of heated materials when $\Delta T_{md}$=20~$^\circ$C}
\label{fig:heatmaterial_20K}
\end{center}
\begin{center}
\includegraphics[width=0.8\columnwidth]{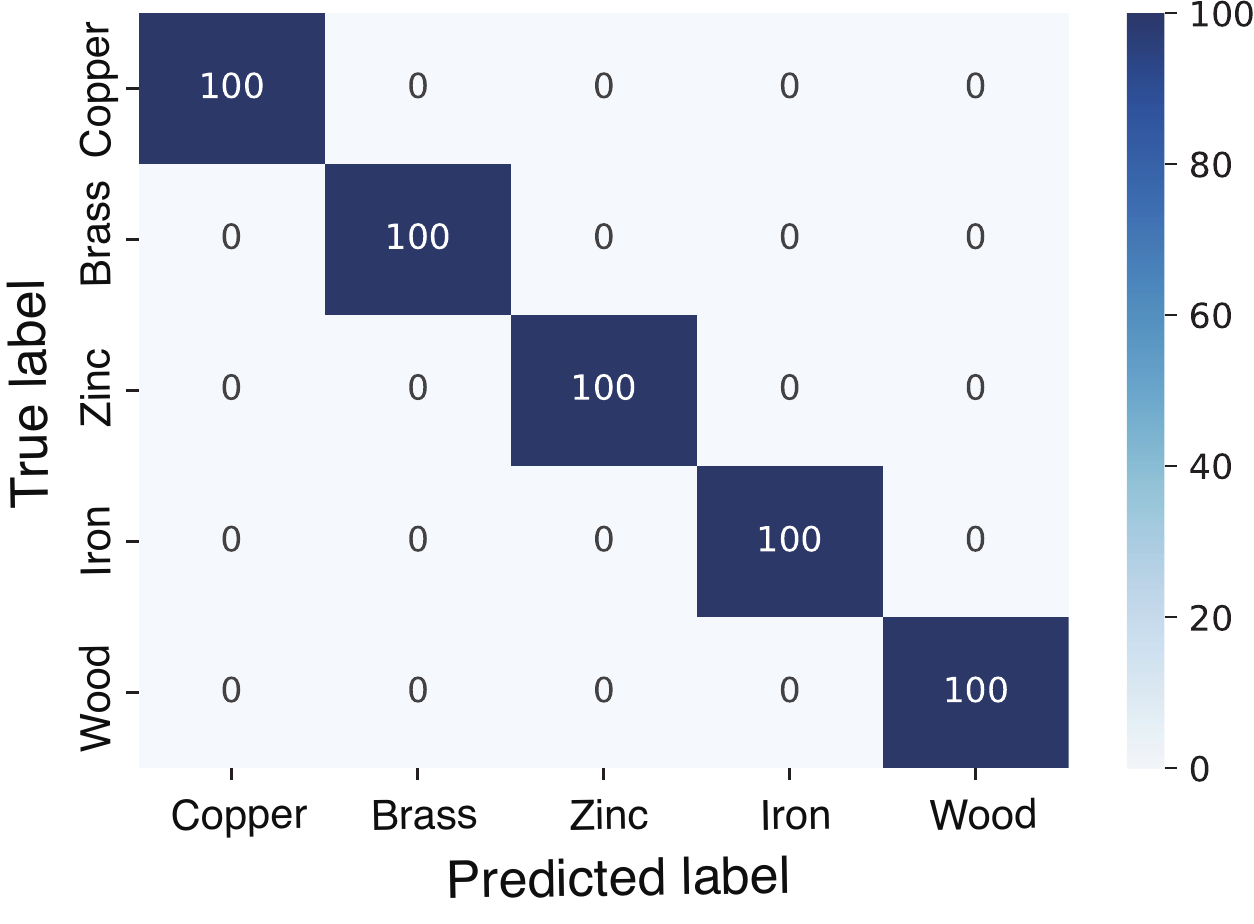}
\caption{[1-A] Heated material classification with cooled gripper.}
\label{fig:CM_heatmaterial_20K}
\end{center}
\end{figure}

\subsection{Experimental results}
\begin{figure}[t!]
\begin{center}
\includegraphics[width=\columnwidth]{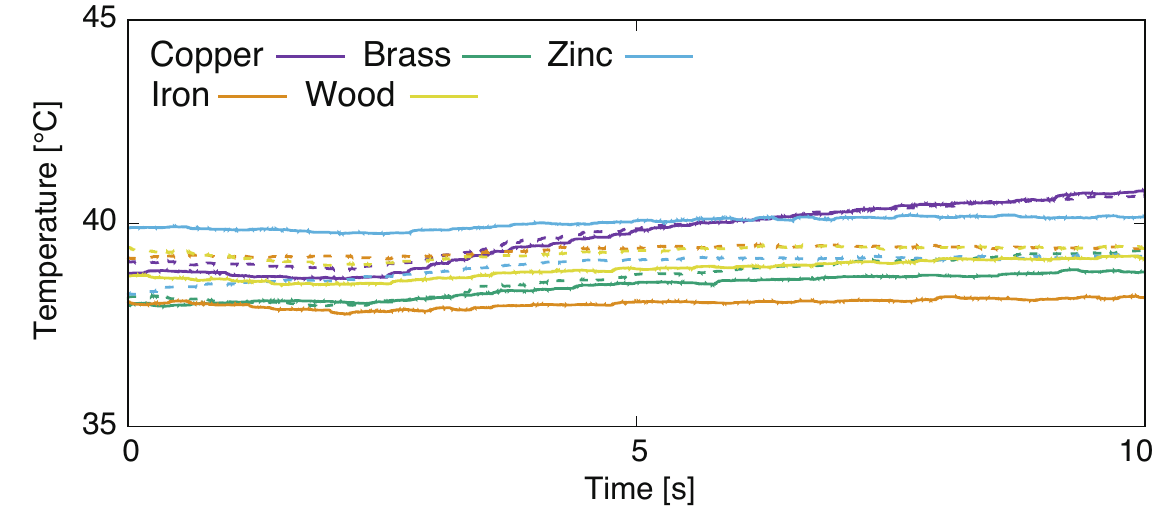}
\caption{[1-B] Test and training data of heated materials when $\Delta T_{md}$=5~$^\circ$C}
\label{fig:heatmaterial_5K}
\end{center}
\begin{center}
\includegraphics[width=0.8\columnwidth]{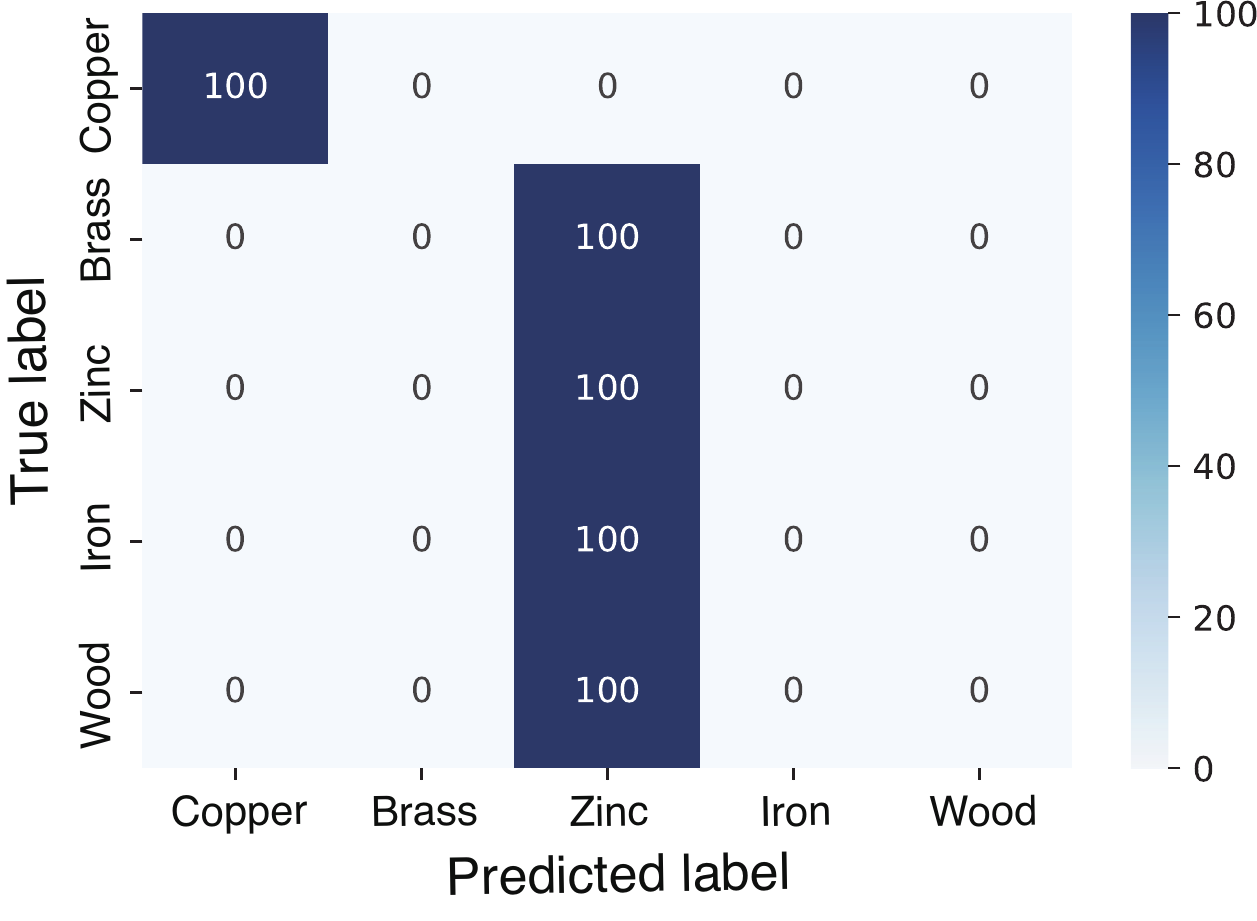}
\caption{[1-B] Heated material classification using the heated gripper.}
\label{fig:CM_heatmaterial_5K}
\end{center}
\end{figure}

Experiment 1 shows the advantage of the proposed system using heated materials. Figs.~\ref{fig:heatmaterial_20K}--\ref{fig:CM_heatmaterial_5K} show the device's temperature responses (10~sec after grasping) used for training (solid line)/test (dotted line) data and their confusion matrices in two cases: [1-A] the robotic gripper is cooled to 18~$^\circ$C (advantage of the proposed gripper) and [1-B] the gripper is heated to 38~$^\circ$C (simulating a conventional heating system). Obviously, the classification accuracy was improved in case [1-A] (Accuracy: 100~$\%$) compared to [1-B] (Accuracy: 40~$\%$); 20~$^\circ$C temperature difference clarifies the differences depending on the materials (Fig.~\ref{fig:heatmaterial_20K}), whereas there is almost no difference as shown in Fig.~\ref{fig:heatmaterial_5K}. 
These results indicate the proposed system can recognize heated materials with higher accuracy than a conventional passive heating system. 

The relationship between the classification accuracy and $\Delta T_{md}$ is investigated in Experiment~2. 
Table~\ref{tab:acc_dT} summarizes the accuracy of material classification when $\Delta T_{md}$ set to [2-A] 5~$^\circ$C, [2-B] 10~$^\circ$C, [2-C] 15~$^\circ$C, and [2-D] 20~$^\circ$C, respectively. From these results, the materials can be completely classified when $\Delta T_{md}$ is more than 10~$^\circ$C. This can be explained by the theoretical model mentioned in Section~II; large $\Delta T_{md}$ reduces the impact of the temperature variation on the classification accuracy.  

\begin{figure*}[t!]
	\begin{minipage}{0.33\hsize}
		\begin{center}
			\includegraphics[width=\hsize,clip]{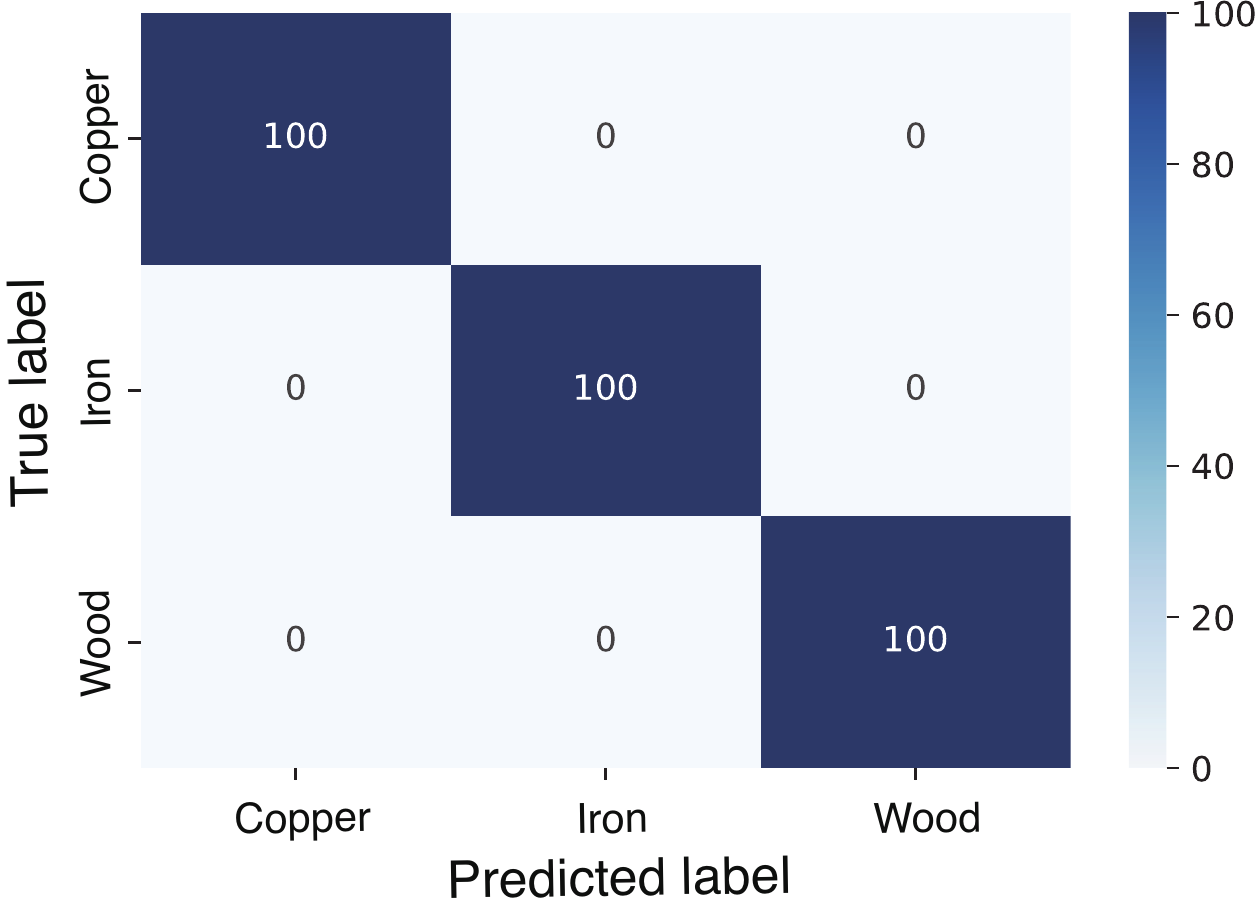}
		\end{center}
	\end{minipage}
	\begin{minipage}{0.33\hsize}
		\begin{center}
			\includegraphics[width=\hsize,clip]{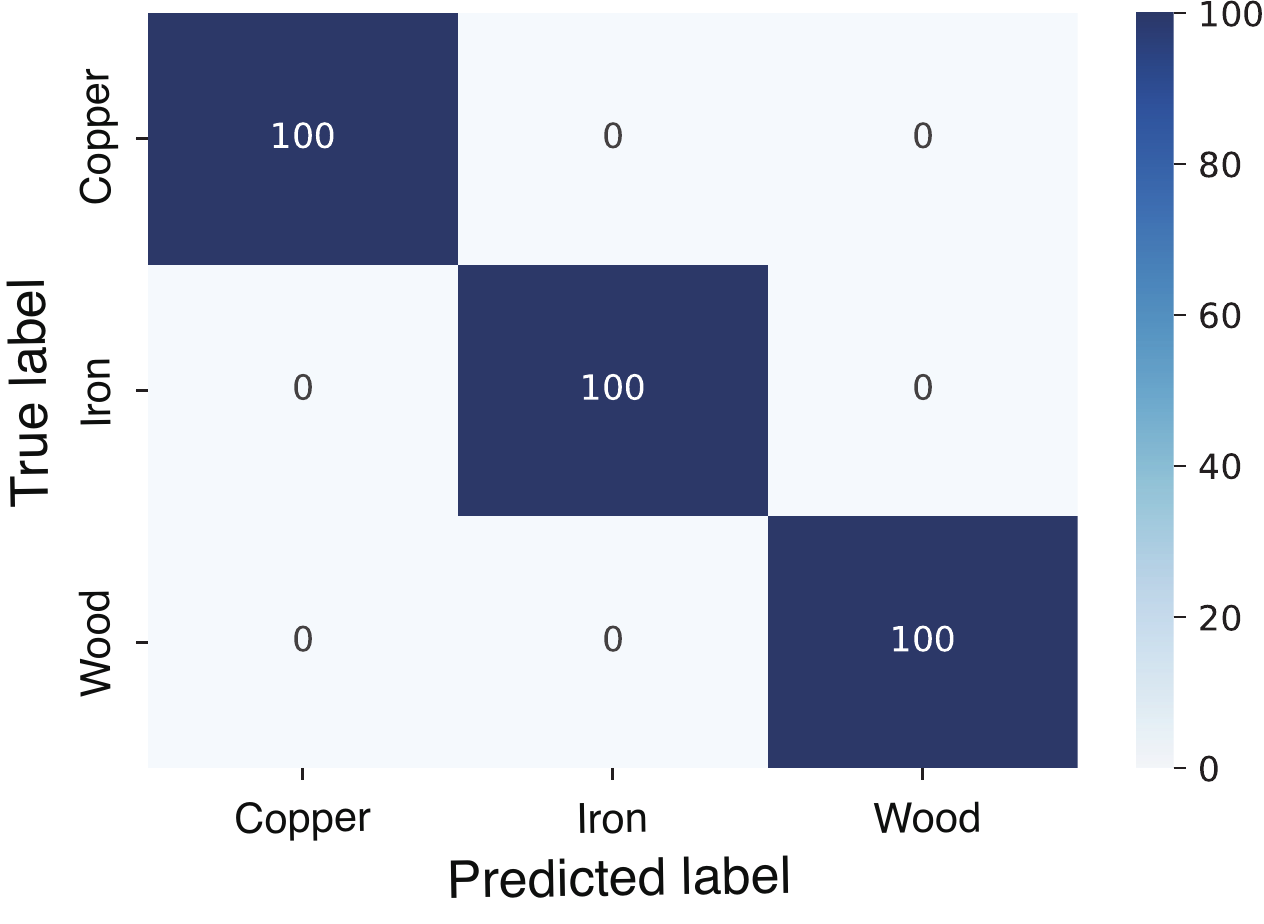}
		\end{center}
	\end{minipage}
	\begin{minipage}{0.33\hsize}
		\begin{center}
			\includegraphics[width=\hsize,clip]{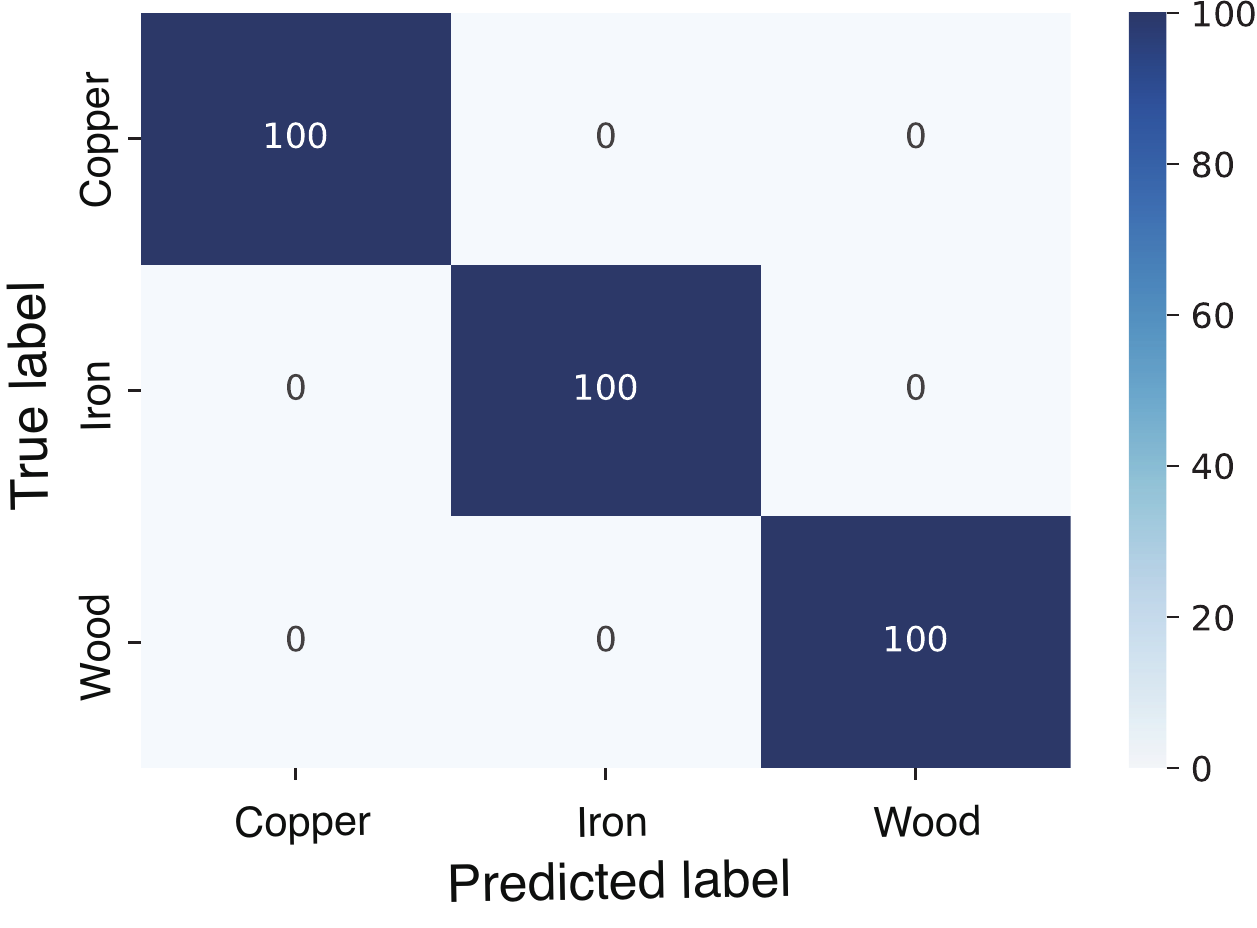}
		\end{center}
	\end{minipage}\\
	\begin{minipage}{0.33\hsize}
		\begin{center}
			\caption{[3-A] Heated material classification.}
			\label{fig:heatmaterial_3M_20K_2}
		\end{center}
	\end{minipage}
	\begin{minipage}{0.33\hsize}
		\begin{center}
			\caption{[3-B] Room-temperature material classification.}
			\label{fig:roommaterial_3M_20K}
		\end{center}
	\end{minipage}
	\begin{minipage}{0.33\hsize}
		\begin{center}
			\caption{[3-C] Cooled material classification.}
			\label{fig:coolmaterial_3M_20K}
		\end{center}
	\end{minipage}
	\end{figure*}

\begin{figure}[t!]
\begin{center}
\includegraphics[width=0.8\columnwidth]{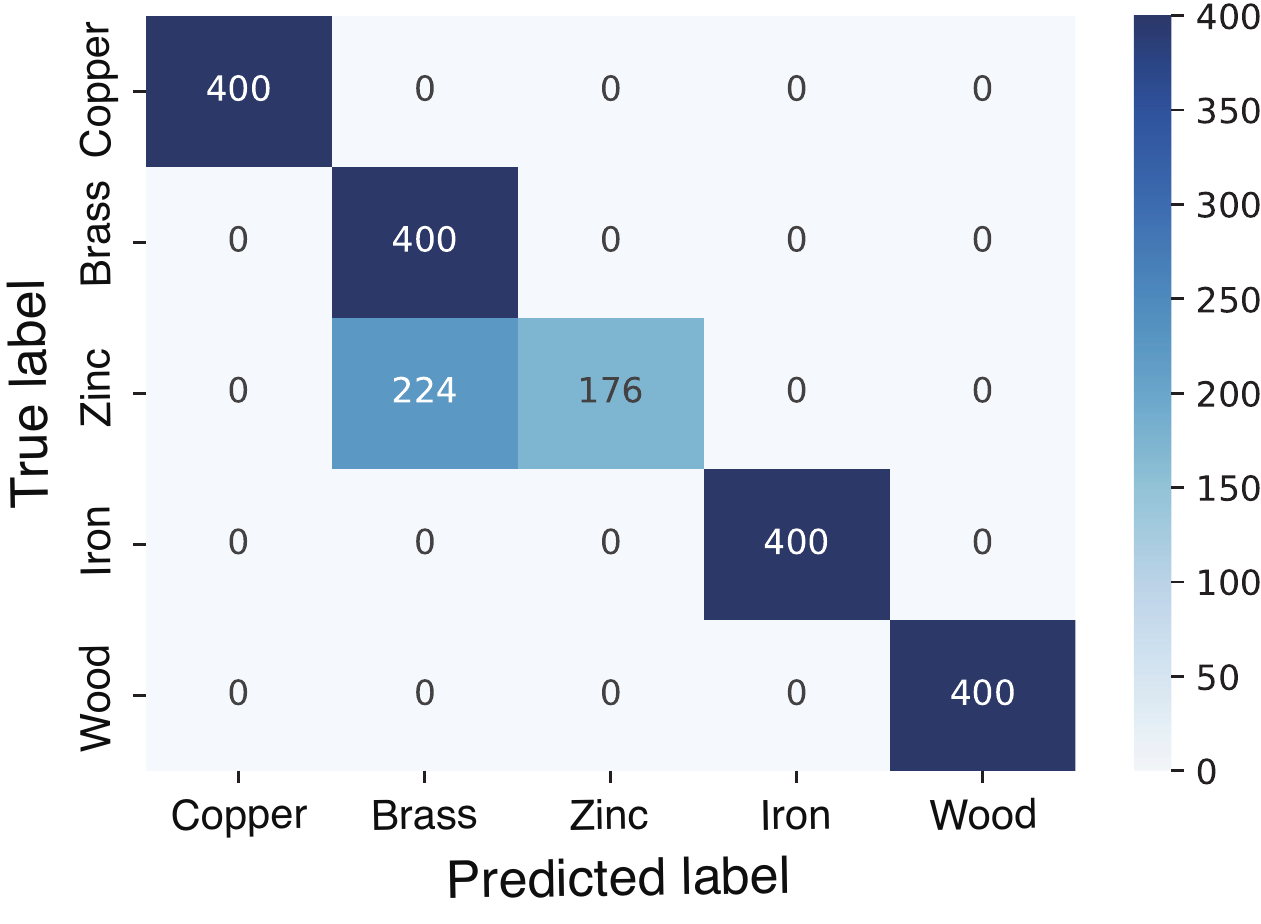}
\caption{Cooled five material classification using the heated gripper.}
\label{fig:CM_coolmaterial_20K}
\end{center}
\end{figure}

In Experiment 3, material classifications in various objects' temperature are conducted so that $\Delta T_{md}$ is 20~$^\circ$C as [3-A] the cooled gripper (23~$^\circ$C) contacted the heated materials (43~$^\circ$C), [3-B] the heated gripper (48~$^\circ$C) contacted the materials at the room temperature (28~$^\circ$C), [3-C] heated gripper (38~$^\circ$C) contacted the cooled materials (18~$^\circ$C). 
The three classification results (copper, iron, and wood) are shown in Figs~\ref{fig:heatmaterial_3M_20K_2} [3-A], \ref{fig:roommaterial_3M_20K} [3-B], and~\ref{fig:coolmaterial_3M_20K} [3-C]. The results show that high accuracy (100$\%$) can be obtained in all cases independent of the object's temperature in the classification. As mentioned in Fig.~\ref{fig:3D_plot}, metal classification is challenging. The accuracy of the heated material classification is high (see Fig.~\ref{fig:CM_heatmaterial_20K}); however, metal similarity causes reduction of the accuracy in case of cooled materials (Fig.~\ref{fig:CM_coolmaterial_20K} shows the classification results using 40 sec data). 
This paper focuses on the temperature difference at the contact surface, yet flow rate and direction of heat flow through the surface can be used to improve the classification accuracy. 

\begin{table}[t!]
		\begin{center}
		\caption{The accuracy of the material classification.} \label{tab:acc_dT}
		\vspace{5pt}
		\begin{tabular}{c|c|c}\hline 
		       Case &$\Delta T_{md}$ [$^\circ$C] & Accuracy [$\%$] \\
			\hline \hline
		      2-A &0 &33.33\\
		      2-B &5 &66.66\\
		      2-C &10 &100\\
		      2-D &15 &100\\
		      2-E &20 &100\\
			\hline 
\end{tabular}
 \end{center}
\end{table}

\section{Conclusion and Future work}
We proposed a material classification system by means of an active-temperature control devise embedded on the robot grasping fingertips. By regulating the initial temperature to be different from the object to be touched, five heated materials (copper, brass, zinc, iron, wood) can be reliably classified; whereas three materials (copper, iron, wood) are classified when heated, cooled, or kept at room temperature. 

There are still possible improvements that we address in future work.
First, the method requires 10~sec temperature time series; however, the recognition time can be substantially shortened using its derivative value by reference to~\cite{bhattacharjee2015rss, bhattacharjee2018ral}. Second, the method needs to measure the object's initial temperature in advance. We can avoid an additional poking step to acquire object temperatures by using cheap infrared cameras or any alternative sensor that outputs in real-time the material temperature to the robotic system. Third, the accuracy of material classification can be further improved, especially for the metals, which can be recognized in combination with vision. Meanwhile, the proposed method can obtain the information of the contact surface while grasping the object; we plan to combine these techniques to improve the recognition accuracy based on the proposed scheme.

\section*{Acknowledgment}
This work was supported in part by Kakenhi Grant-in-Aid for Young Scientists Number 21K21325. 
The authors thank Dr.~Toshio Ueshiba from AIST for implementing ROS on the Academic SCARA robot.

\bibliographystyle{IEEEtran}
\bibliography{reference}

\end{document}